\documentclass[11pt,a4paper]{article}
\usepackage[hyperref]{emnlp-ijcnlp-2019}
\usepackage{times}
\usepackage{latexsym}

\usepackage{url}

\aclfinalcopy 

\usepackage{natbib}
\usepackage{graphicx}
\usepackage{hyperref}
\usepackage{siunitx}
\usepackage{booktabs}
\usepackage{longtable}
\usepackage{caption}
\usepackage[]{algorithm2e}
\usepackage[normalem]{ulem}
\usepackage{float}

\usepackage{subcaption}

\usepackage{enumitem}
\usepackage{amsfonts}
\usepackage{chngcntr}

\usepackage{amsmath}
\usepackage{bbold}

\DeclareMathOperator*{\E}{\mathbb{E}}

\newcommand{\com}[1]{}	
\newcommand{\camready}[1]{}

\com{
\newcommand\floatc@simplerule[2]{{\@fs@cfont #1 #2}\par}
\newcommand\fs@simplerule{\def\@fs@cfont{\bfseries}\let\@fs@capt\floatc@simplerule
  \def\@fs@pre{\hrule height.8pt depth0pt \kern4pt}%
  \def\@fs@mid{\kern4pt\hrule height.8pt depth0pt \kern4pt \relax}%
  \def\@fs@post{}%
  \let\@fs@iftopcapt\iffalse}  
}

\newif{\ifhidecomments}

\ifhidecomments
    \newcommand{\jdcomment}[1]{}	
    \newcommand{\nascomment}[1]{}
    \newcommand{\dallas}[1]{}
    \newcommand{\roy}[1]{}
    \newcommand{\suchin}[1]{}
    \newcommand{\roys}[1]{}
    \newcommand{\resolved}[1]{}
\else
    \newcommand{\jdcomment}[1]{\textcolor{teal}{[#1 ({\bf Jesse})]}} 
    \newcommand{\nascomment}[1]{\textcolor{red}{[#1 ({\bf Noah})]}} \newcommand{\dallas}[1]{\textcolor{orange}{[#1 ({\bf DBC})]}} 
    \newcommand{\suchin}[1]{\textcolor{cyan}{[#1 ({\bf SG})]}} 
    \newcommand{\roy}[1]{\textcolor{blue}{[#1 ({\bf Roy})]}} 
    \newcommand{\ana}[1]{\textcolor{purple}{[#1 ({\bf Ana})]}} 
    \newcommand{\roys}[1]{\textcolor{blue}{\sout{#1}}} 
    \newcommand{\resolved}[1]{} 
\fi
 
\title{Show Your Work:  Improved Reporting of Experimental Results}

\author{Jesse Dodge$^\clubsuit$ \quad Suchin Gururangan$^\diamondsuit$  \quad 
Dallas Card$^\heartsuit$ \quad 
Roy Schwartz$^\spadesuit$$^\diamondsuit$ \quad
  Noah A. Smith$^\spadesuit$$^\diamondsuit$ \\
  $^\clubsuit$Language Technologies Institute,
  Carnegie Mellon University, Pittsburgh, PA, USA\\ $^\diamondsuit$Allen Institute for Artificial Intelligence, Seattle, WA, USA \\
  $^\heartsuit$Machine Learning Department, Carnegie Mellon
  University, Pittsburgh, PA, USA \\
  $^\spadesuit$Paul G. Allen School of Computer Science \& Engineering,
  University of Washington, Seattle, WA, USA \\
  {\tt \{jessed,dcard\}@cs.cmu.edu \quad \{suching,roys,noah\}@allenai.org}}

\date{}

\begin{document}

\maketitle
\begin{abstract}
Research in natural language processing proceeds, in part, by demonstrating that new models achieve superior performance (e.g., accuracy) on held-out test data, compared to previous results. 
In this paper, we demonstrate that test-set performance scores alone are insufficient for drawing accurate conclusions about which model performs best.
We argue for reporting additional details, especially performance on validation data  obtained during model development.
We present a novel technique for doing so: \emph{expected validation performance} of the best-found model as a function of computation budget (i.e., the number of hyperparameter search trials or the overall training time\resolved{ \dallas{should make this i.e. what we want people to report -- trials, hours, \$, etc.}}).
Using our approach, we find multiple recent model comparisons where authors would have reached a different conclusion if they had used more (or less) computation.
Our approach also allows us to estimate the amount of computation required to obtain a given accuracy; applying it to several recently published results yields massive variation across papers, from hours to weeks.
We conclude with a set of best practices for reporting experimental results which allow for robust future comparisons, and provide code to allow researchers to use our technique\com{ with minimal additional effort}.\footnote{\url{https://github.com/allenai/allentune}}

\end{abstract}

\section{Introduction}

\resolved{\jdcomment{Things to add: citation to green ai, citation to strubell et al}}

\begin{figure}[!t]
\centering


\com{\caption*{\medium \textbf{Expected performance for given budget}}}
\centering
\hspace*{-10mm}
\includegraphics[scale=0.33,clip,trim={6.5cm .6cm 4cm 1.2cm }]{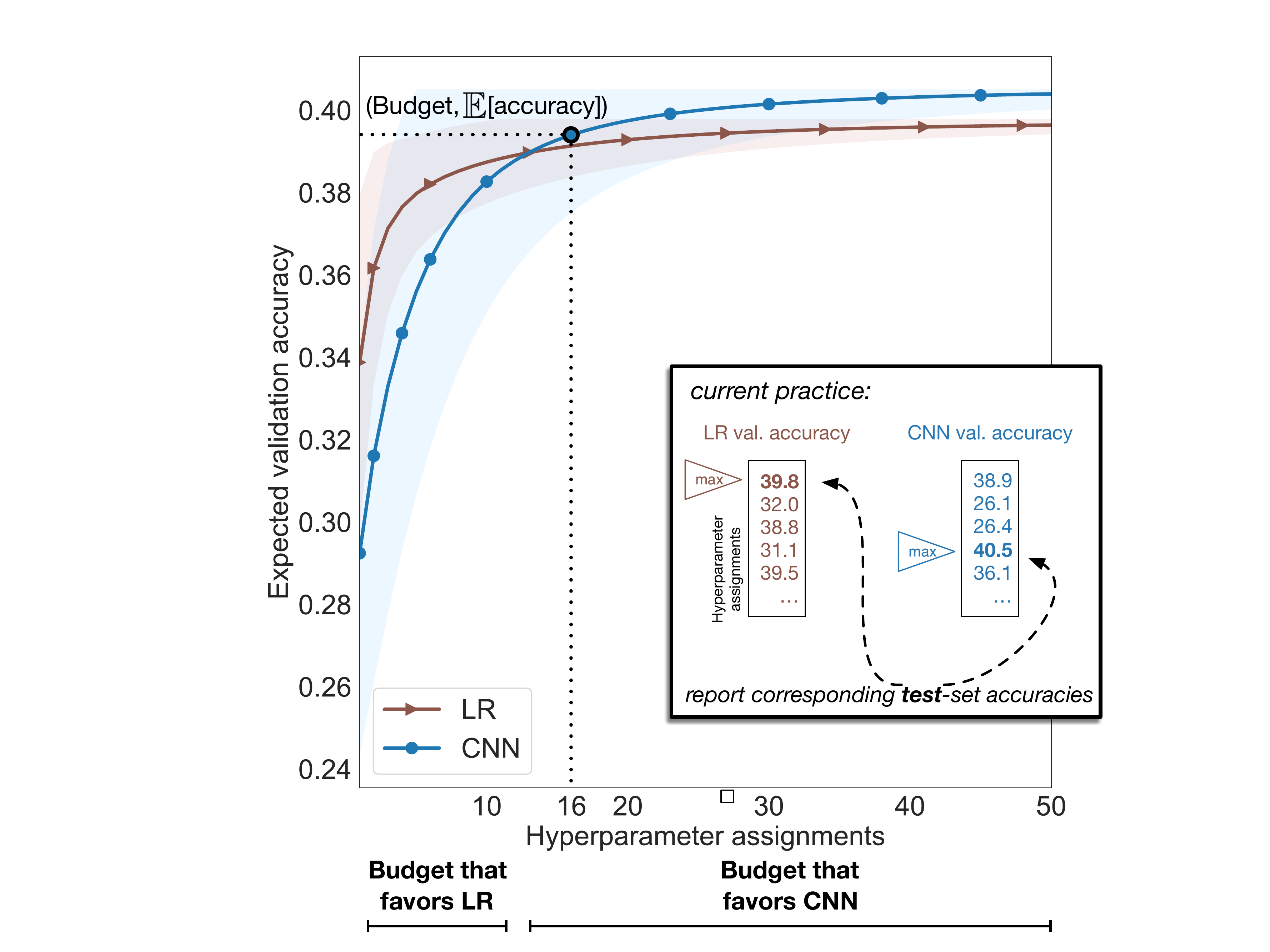}

\caption{\label{fig:simple_models_SST (fine-grained)}
Current practice when comparing NLP models is to train multiple instantiations of each, choose the best model of each type based on validation performance, and compare their performance on test data (inner box).
Under this setup, (assuming test-set 
results are similar to validation),
one would conclude from the results above (hyperparameter search for two models on the 5-way SST classification task) that the CNN outperforms Logistic Regression (LR).
In our proposed evaluation framework, we instead encourage practitioners to consider the expected validation accuracy ($y$-axis; shading shows $\pm 1$ standard deviation), as a function of budget ($x$-axis).
Each point on a curve is the \emph{expected value} of the best validation accuracy obtained ($y$) after evaluating $x$ random hyperparameter values.
\com{These curves represent the expected value of the accuracy that would be achieved by the best single model obtained using the corresponding budget for hyperparameter tuning.
Results on the fiveway SST dataset for logistic regression and a CNN.
We give the hyperparameter search space in Appendix~\ref{app:simple_models_app}.
We sample 50 hyperparameter assignments for each model, and display variance bands as a shaded regions.} 
Note that (1) the better performing model depends on the computational budget; LR has higher expected performance for budgets up to 10 hyperparameter assignments, while the CNN is better for larger budgets.
(2) Given a model and desired accuracy (e.g., 0.395 for CNN), we can
estimate the expected budget required to reach it (16; dotted
lines). 
\resolved{\nascomment{the numbers are different now, please change 0.4
and 23 appropriately}}
\resolved{\roy{Suchin, can you please add value to the X axis as well for the dotted line?}}
\resolved{\nascomment{maybe also talk about variance; perhaps we would choose the CNN at $B = 10$ anyway if we're feeling lucky about variance?}}
}
\end{figure}

\resolved{\roy{Change style file to emnlp 2019}}

\resolved{\jdcomment{We could say in the abstract that we improve over recent work, by having an exact version of something they approximate.\\
add some numbers to the abstract\\
} }

\resolved{\jdcomment{Noah: Is it unclear that we are encouraging readers to
  generate and report expected performance curves? A possible other
  reading is that we are encouraging reporting just mean and standard
  deviation, and we show curves to indicate why that's important.}
\nascomment{resolved below, maybe?}}
In NLP and machine learning, improved performance on held-out test data is typically used as an indication of the superiority of one method over others.
But, as the field grows, there is an increasing gap between the large computational budgets used for some high-profile experiments and the budgets used in most other work \cite{GreenAI}. This hinders meaningful comparison between experiments, as improvements in performance can, in some cases, be obtained purely through more intensive hyperparameter tuning \citep{melis2018, lipton2018}.\footnote{\com{While the results of such efforts are valuable, they also lead to environmental and social tolls}Recent work has also called attention to the environmental cost of intensive model exploration \cite{energy}. }\com{; it is hard to have a fair comparison between approaches when they have dramatically different experimentation budgets (or when the budgets are simply not reported)}

Moreover, recent investigations into ``state-of-the-art'' claims have found competing methods to only be comparable, without clear superiority, even against baselines \cite{score_distributions, lucic2018, li2019}\resolved{\suchin{nitpick, but lucic et al 2018 is about GANs and vision / Li 2019 evaluates on CIFAR (ie, not NLP). Should we open this paragraph with "In machine learning", instead of "in NLP", and then in the next paragraph say that we are echoing these papers but focus on NLP specifically?}}; 
this has exposed the need for reporting more than a single point estimate of performance.

\com{Third, providing only the performance of the best model on test data creates a perverse inventive to treat the test set as leaderboard rather than unbiased evaluation data.} 

Echoing calls for more rigorous scientific practice in machine learning \cite{lipton2018,sculley2018}, we draw attention to the weaknesses in current reporting practices and propose solutions which would allow for fairer comparisons and improved reproducibility.

\resolved{\nascomment{I worked over the remainder of the intro and the caption for the figure; this part looks good}}

Our primary technical contribution is the introduction of a tool for reporting validation results in an easily interpretable way: \textit{expected validation performance}  of the best model under a given computational budget.\footnote{We use the term \emph{performance} as a general evaluation measure, e.g., accuracy, $F_1$, etc.}
That is, given a budget sufficient for training and evaluating $n$ models, we calculate the expected performance of the best of these models on validation data.  Note that this differs from the \emph{best observed} value after $n$ evaluations.
Because the expectation can be estimated from the distribution of $N$
validation performance values, with $N\geq n$, and these are obtained during model
development,\footnote{We
  leave forecasting performance with larger budgets $n>N$ to future
  work.}  our method \textbf{does not require additional computation}
beyond hyperparameter search or optimization. 
\resolved{\nascomment{reworded a
  bit here to emphasize ... check}}  
We encourage researchers to report
expected validation performance as a curve, across values of $n \in
\{1, \ldots, N\}$. \resolved{\nascomment{added last sentence; check flow and redundancy}}

As we show in \S\ref{sec:contextual_embeddings}, our approach makes clear that the expected-best performing model is a function of the computational budget.
In \S\ref{sec:predict_compute} we show how our approach can be used to estimate the budget that went into obtaining previous results; in one example, we see a too-small budget for baselines, while in another we estimate a budget of about 18 GPU days was used (but not reported).
Previous work on reporting validation performance used the bootstrap to approximate the mean and variance of the best performing model \cite{lucic2018}; in \S\ref{sec:bootstrap} we show that our approach computes these values with strictly less error than the bootstrap.\resolved{\roy{We don't discuss our empirical results anywhere}}

We conclude by presenting a set of recommendations for researchers that will improve scientific reporting over
current practice.
 We emphasize this work is about \emph{reporting}, not about running additional experiments (which undoubtedly can improve evidence in comparisons among models).
 Our reporting recommendations 
aim at reproducibility and improved understanding of sensitivity to hyperparameters and random initializations.
Some of our recommendations may seem obvious; however, our empirical analysis shows that out of fifty EMNLP 2018 papers chosen at random, none report all items we suggest.

\section{Background}

\paragraph{Reproducibility}
Reproducibility in machine learning is often defined as the ability to produce the \emph{exact} same results as reported by the developers of the model.
In this work, we follow \citet{gundersen2018} and use an extended notion of this concept: when comparing two methods, two research groups with different implementations should follow an experimental procedure which leads to the same conclusion about which performs better. 
As illustrated in Fig.~\ref{fig:simple_models_SST (fine-grained)}, this conclusion often depends on the amount of computation applied. 
Thus, to make a \emph{reproducible} claim about which model performs best, we must also take into account the budget used (e.g., the number of hyperparameter trials).

\paragraph{Notation}
We use the term \emph{model family}\com{ (denoted $\mathcal{M}$\resolved{ \nascomment{do we really need that?}})} to refer to an approach subject to comparison and to hyperparameter selection.\com{Standard practice selects a single model per family, using validation (also called ``development'') data performance, reporting performance on a held-out test set.}\footnote{Examples include different architectures, but also ablations of the same architecture.}
Each model family $\mathcal{M}$ requires its own hyperparameter selection, in terms of a set of $k$ hypermarameters, each of which defines a range of possible values. 
A \emph{hyperparameter value} (denoted $h$) is a $k$-tuple of specific values for each hyperparameter.
We call the set of all possible hyperparameter values $\mathcal{H}_{\mathcal{M}}$.\footnote{The hyperparameter value space can also include the random seed used to initialize the model, and some specifications such as the size of the hidden layers in a neural network, in addition to commonly tuned values such as learning rate.}  
Given $\mathcal{H}_{\mathcal{M}}$ and a computational budget sufficient for training $B$ models, the set of hyperparameter values is $\{ h_1, \ldots, h_B\}$, $h_i \in \mathcal{H}_{\mathcal{M}}$. 
We let $m_i \in \mathcal{M}$ denote the model trained with hyperparameter value $h_i$.

\resolved{\ana{Hyperparameter values are a k-tuple $\Rightarrow$ is a k-tuple?}\\
\ana{Suggestion: Add a symbol $h$ after introducing hyperparameter values}}

\paragraph{Hyperparameter value selection}
There are many ways of selecting hyperparameter values, $h_i$.
Grid search and uniform sampling\com{ (i.e., uniformly from $\mathcal{H}_{\mathcal{M}}$)} are popular systematic methods; the latter has been shown to be superior for most search spaces \citep{bergstra2012}.
Adaptive search strategies such as Bayesian optimization select $h_i$ after evaluating $h_1,\ldots,h_{i-1}$.   While these strategies may find better results quickly, they are generally less reproducible and harder to parallelize \cite{hyperband}.
Manual search, where practitioners use knowledge derived from previous experience to adjust hyperparameters after each experiment, is a type of adaptive search that is the least reproducible, as different practitioners make different decisions.
Regardless of the strategy adopted, we advocate for detailed reporting of the method used for hyperparmeter value selection\camready{ and the budget} (\S\ref{sec:recommendations}).
We next introduce a technique to visualize results of samples which are drawn i.i.d.~(e.g., random initializations or uniformly sampled hyperparameter values).

\section{Expected Validation Performance Given Budget} \label{sec:method}

\resolved{\suchin{There's a pretty large gap between Figure 1 and 2, can we display graphics in Section 3? I think someone had mentioned a plot around bootstrap comparison}}

\resolved{\roy{The beginning here feels like part of the background (belongs to the same sections as \emph{Definitions} above). I would start this with some high level discussion.
Reminding the reader again of our motivation ($n$ is never reported, but important...), and then explicitly present our main contribution: the expected max equation.}}

\resolved{\dallas{An alternative way of starting this section: given the results of a hyperparameter search, it is common for NLP for researchers to keep the best model found, evaluate it on test data, and report that number. However, there is useful information in the intermediate evaluations on validation data that we propose to make use of.}}

After selecting the best hyperparameter values $h_{i^\ast}$ from among $\{ h_1, \ldots, h_B\}$ with actual budget $B$, NLP researchers typically evaluate the associated model $m_{i^\ast}$ on the test set and report its performance as an estimate of the family $\mathcal{M}$'s ability to generalize to new data.  We propose to make better use of the intermediately-trained models $m_1, \ldots, m_B$.


\resolved{\nascomment{changed here} }For any set of $n$ hyperparmeter values, denote the validation performance of the best model as
\begin{equation}
    v_n^\ast = \textstyle \max_{h\in \{h_1, \ldots, h_n\}} \mathcal{A}(\mathcal{M}, h, \mathcal{D}_T, \mathcal{D}_V), \label{eq:max}
\end{equation}
where $\mathcal{A}$ denotes an algorithm that returns the performance on validation data $\mathcal{D}_V$ after training a model from family $\mathcal{M}$ with hyperparameter values $h$ on
training data $\mathcal{D}_T$.\footnote{$\mathcal{A}$ captures standard parameter estimation, as well as procedures that depend on validation data, like early stopping.}
We view evaluations of $\mathcal{A}$ as the elementary unit of experimental cost.\footnote{Note that researchers do not always report validation, but rather \emph{test} performance, a point we will return to in \S\ref{sec:recommendations}\resolved{\nascomment{check that we do}}.}


Though not often done in practice, procedure \eqref{eq:max} could be repeated many times with different hyperparameter values, yielding a \emph{distribution} of values for random variable $V_n^\ast$. \resolved{\ana{If I understood correctly that you’re suggesting  repeating (1) with different sequence of HPs values, doesn't the number of HPs configurations become bigger (number of times that you repeat * n)? I think it is not clear to be what the procedure refers to exactly and why it could be repeated without additional cost.} }This would allow us to estimate the \emph{expected} performance, $\E[V^\ast_n \mid n]$ (given $n$ hyperparameter configurations).  The key insight used below is that, if we use random search for hyperparameter selection, then the effort that goes into a single round of random search (Eq.~\ref{eq:max}) suffices to construct a useful estimate of expected validation performance,
without requiring \emph{any further experimentation}.

Under random search, the $n$ hyperparameter values $h_1, \ldots, h_n$ are drawn uniformly at random from $\mathcal{H}_{\mathcal{M}}$, so the values of $\mathcal{A}(\mathcal{M}, h_i, \mathcal{D}_T, \mathcal{D}_V)$ are i.i.d.
As a result, the maximum among these is itself a random variable. We introduce a diagnostic that captures information about the computation used to generate a result: the expectation of maximum performance, \emph{conditioned} on $n$, the amount of computation used in the maximization over hyperparameters and random initializations:
\begin{equation}  \label{eq:expected_max}
    \E \left[\textstyle \max_{h\in \{ h_1, \ldots, h_n\}} \mathcal{A}(\mathcal{M}, h, \mathcal{D}_T, \mathcal{D}_V) \mid n  \right].
\end{equation}

Reporting this expectation as we vary $n \in \{1, 2, \ldots, B\}$ gives more information than the maximum $v_B^\ast$ (Eq.~\ref{eq:max} with $n=B$); future researchers who use this model will know more about the computation budget  required to achieve a given performance.
We turn to calculating this expectation, then we 
compare it to the bootstrap (\S\ref{sec:bootstrap}), and discuss estimating variance (\S\ref{sec:variance}).\resolved{\roy{todo: change intuitions to a different name? discuss other subsections} }
\resolved{\nascomment{is it the variance of the estimator for the expectation we discuss (what this text seems to say) or an estimate of $V^\ast$'s variance (I think the latter)?}}

\subsection{Expected Maximum}\label{sec:expected_max}
\resolved{\roy{This is generally good, but you need to be a bit more explicit in walking the reader through the procedure. Why are we doing this? what is result of this process?}}

We describe how to estimate the expected maximum validation performance (Eq.~\ref{eq:expected_max}) given a budget of $n$ hyperparameter values.\footnote{Conversion to alternate formulations of budget, such as GPU hours or cloud-machine rental cost in dollars, is straightforward in most cases.}

Assume we draw $\{ h_1, \ldots, h_n\}$ uniformly at random from hyperparameter space $\mathcal{H}_{\mathcal{M}}$.
Each evaluation of $\mathcal{A}(\mathcal{M}, h, \mathcal{D}_T, \mathcal{D}_V)$ is therefore an i.i.d.~draw of a random variable, denoted $V_i$, with observed value $v_i$ for $h_i \sim \mathcal{H}_\mathcal{M}$.  Let the maximum among $n$ i.i.d.~draws from an unknown distribution be
\begin{align}
V_n^\ast & = \textstyle \max_{i \in \{1,\ldots,n\}}  V_i
\end{align}
We seek the expected value of $V^*_n$ given $n$:
\begin{align}\label{eq:expect_as_sum}
\E[V^\ast_n \mid n] &= \textstyle \sum_v  v\cdot P(V^\ast_n = v \mid n) 
\end{align}
where $P(V^\ast_n \mid n)$ is the probability mass function (PMF) for the max-random variable.\footnote{For a finite validation set $\mathcal{D}_V$, most performance measures (e.g., accuracy) only take on a finite number of possible values, hence the use of a sum instead of an integral in Eq.~\ref{eq:expect_as_sum}.} 
For discrete random variables, 
\begin{align}\label{eq:discrete_pdf}
P(V_n^* = v\mid n) = P(V_n^* \leq v\mid n) - P(V_n^* < v\mid n),
\end{align}

Using the definition of ``max'', and the fact that the $V_i$ are drawn i.i.d.\com{ (still suppressing the conditioning on $n$)}, 
\begin{align}
    P(V_n^* &\leq v\mid n) = P\left( \textstyle\max_{i\in\{1,...,n\}} V_i  \leq v\mid n\right) \nonumber \\
    &= P(V_1 \leq v, V_2 \leq v, \ldots, V_n \leq v\mid n) \nonumber \\
    &= \textstyle \prod_{i=1}^n P(V_i \leq v)  = P(V \leq v)^n,
\end{align}
and similarly for $P(V_n^* < v\mid n)$.

$P(V \leq v)$ and $P(V < v)$ are cumulative distribution functions, which we can estimate using the empirical distribution, i.e.
\begin{align}
    \hat P(V \leq v) = \textstyle \frac{1}{n} \sum_{i=1}^n \mathbb{1}_{[V_i \leq v]}
\end{align}
and similarly for strict inequality.
\resolved{\ana{Isn't $\mathbb{1}_{[V_i \leq v]}$ more standard for the characteristic function? How did you bring the conditioning back?}}

Thus, our estimate of the expected maximum validation performance is 
\begin{equation}
\hat \E[V^\ast_n \mid n] = \textstyle\sum_{v}  v \cdot ( \hat P(V_i \leq v)^n - \hat P(V_i < v)^n).
\end{equation}
\resolved{\roy{@jesse, I brought back the conditioning on $n$ because I felt it was too confusing w/o it. Can you please double check that I didn't miss anything?}}

\paragraph{Discussion}
As we increase the amount of computation for evaluating hyperparameter values ($n$), the maximum among the samples will approach the observed maximum $v^\ast_B$\resolved{ \nascomment{check}}.  Hence the curve of $\E[V^\ast_n \mid n]$ as a function of $n$ will appear to asymptote.  Our focus here is not on estimating that value, and we do not make any claims about extrapolation of $V^\ast$ beyond $B$, the number of hyperparameter values to which $\mathcal{A}$ is actually applied.

Two points follow immediately from our derivation.  First, at $n=1$, $\E[V^\ast_1 \mid n=1]$ is the mean of $v_1, \ldots, v_n$.  Second, for all $n$, $\E[V^\ast_n \mid n] \le v_n^\ast = \max_{i} v_i$, which means the curve is a lower bound on the selected model's validation performance.

\subsection{Comparison with Bootstrap}\label{sec:bootstrap}
\resolved{\nascomment{make this punchier}}

\citet{lucic2018} and \citet{henderson2018} have advocated for using the bootstrap to estimate the mean and variance of the best validation performance. The bootstrap \citep{bootstrap} is a general method which can be used to estimate statistics that do not have a closed form.
The bootstrap process is as follows: draw $N$ i.i.d. samples (in our case, $N$ model evaluations). From these $N$ points, sample $n$ points (with replacement), and compute the statistic of interest (e.g., the max).
Do this $K$ times (where $K$ is large), and average the computed statistic. By the law of large numbers, as $K\to \infty$ this average converges to the sample expected value \cite{bootstrap}. 
\resolved{\jdcomment{this is correct when $n=N$, but when $n<N$ it converges to the expected value of $n$ given the $N$ something.}}
\resolved{\nascomment{I am slightly worried about $n$, $N$, and $B$ being used consistently throughout}
\jdcomment{it converges to the expected value we compute. }}

The bootstrap has two sources of error: the error from the finite sample of $N$ points, and the error introduced by resampling these points $K$ times. 
Our approach has strictly less error than using the bootstrap: our calculation of the expected maximum performance in \S\ref{sec:expected_max} provides a closed-form solution, and thus contains none of the resampling error (the finite sample error is the same).



\resolved{\nascomment{we said in an earlier draft that our model computes $\E[V^\ast_n]$ exactly; that's not right, because we are using the empirical cdf ... so I'm not sure what to say here.  old text commented out below}
\jdcomment{we were trying to say that the bootstrap is an approximation specifically to our approach (which itself is an approximation to $\E[V^\ast_n]$). the second sentence commented out below does a better job of getting at the idea: the variance of the bootstrapped estimator converges to the standard error of the full sample, which we can compute exactly (without the resampling error introduced by the bootstrap).\\
there are two types of error at play when using the bootstrap: the sampling error introduced by having a finite sample $N$ from the population, and the resampling error introduced by the bootstrap's sampling $n$ (with replacement, many times).\\
the bootstrap proceeds as follows:\\
From the original sample of $N$ points, sample $n$ points (with replacement). Then, compute the statistic of interest (in our case, the max).\\
Do this $K$ times (where $K$ is large), and average the results.\\
This average, as $K$ goes to infinity, converges to our estimate of $\E[V^\ast_n]$. \\
Note that our estimate of $\E[V^\ast_n]$ still has the sampling error introduced by having a finite sample $N$ from the population, but it has none of the resampling error from the bootstrap.\\
}}



\subsection{Variance of $V^\ast_n$}\label{sec:variance}
\resolved{\nascomment{worked over this, please check}}

Expected performance becomes more useful with an estimate of variation.
When using the bootstrap, standard practice is to report the standard deviation of the estimates from the $K$ resamples. 
As $K\to \infty$, this standard deviation approximates the sample standard error \cite{bootstrap}.
We instead calculate this from the distribution in Eq.~\ref{eq:discrete_pdf} using the standard plug-in-estimator. \resolved{\nascomment{check that. I don't think we need to give a formula for this.}}

In most cases, we advocate for reporting a measure of variability such as the standard deviation or variance; however, in some cases it might cause confusion. 
For example, when the variance is large, plotting the expected value plus the variance can go outside of reasonable bounds, such as accuracy greater than any observed (even greater than 1).
\resolved{For example, with a small computation budget for computationally expensive problems, the expected value might be close to random performance, with a large variance.\roy{do we have an example in future section? e.g., BERT on STILTS or sthing?}
The variance bars on the low end can then go well below random chance (potentially even going negative), which can be quite misleading.}
In such situations, we recommend shading only values within the observed range, such as in Fig.~\ref{fig:squad}.
\resolved{\roy{not sure this reference helps. looking at Fig.~\ref{fig:squad} (which is a few pages later), I don't see the intuition. Consider dropping it}}
Additionally, in situations where the variance is high and variance bands overlap between model families (e.g., Fig.~\ref{fig:simple_models_SST (fine-grained)}), the mean is still the most informative statistic.

\section{Case Studies}
Here we show two clear use cases of our method.
First, we can directly estimate, for a given budget, which approach has better performance. 
Second, we can estimate, given our experimental setup, the budget for which the reported validation performance ($V^\ast$) matches a desired performance level.
We present three examples that demonstrate these use cases. 
First, we reproduce previous findings that compared different models for text classification.
Second, we explore the time vs.~performance tradeoff of models that use contextual word embeddings \citep{Peters2018DeepCW}.
Third, from two previously published papers, we examine the budget required for our expected performance to match their reported performance.
\com{the estimate the budget required to match the reported performance.\com{and ask
``for what budget does the reported validation performance ($V^\ast$) match the expected performance, given our experimental setup?"}}
We find these budget estimates vary drastically.
Consistently,  we see that the best model is a function of the budget.
\resolved{\dallas{How do we know if this is correct?} \suchin{yeah, I think we need to be careful about what exactly we want to show with the "looking back" plots. we don't \emph{really} know how much compute they applied, but we can make a rough estimate.}} We publicly release the search space and training configurations used for each case study. \footnote{\url{https://github.com/allenai/show-your-work}}

\resolved{
\suchin{We need to be able to describe the time-based plots succinctly. How about annotating the highest expected max in each plot? \\
it's clear that these many of these models/tasks came out of AI2, should we should mention that we chose models based on ability to reproduce with the same library (ie AllenNLP, but maybe not explicitly saying this).\\}}

\resolved{\dallas{I wonder if we need to say something in here about test numbers, even just to say we're not reporting them because...} \nascomment{yes, good idea} \dallas{my attempt:}}
Note that we do not report test performance in our experiments, as our purpose is not to establish a benchmark level for a model, but to demonstrate the utility of expected validation performance for model comparison and reproducibility.

\subsection{Experimental Details}

For each experiment, we document the hyperparameter search space,
hardware, average runtime, number of samples, and links to model implementations.
We use public implementations for all models in our experiments, primarily in AllenNLP \citep{Gardner2018AllenNLPAD}\com{ and scikit-learn}.
We use Tune \citep{liaw2018tune} to run parallel evaluations of uniformly sampled hyperparameter values. 


\subsection{Validating Previous Findings}\label{sec:simple_models}
We start by applying our technique on a text classification task in order to confirm a well-established observation \citep{Yogatama2015BayesianOO}: logistic regression has reasonable performance with minimal hyperparameter tuning, but a well-tuned convolutional neural network (CNN) can perform better. 

We experiment with the fine-grained Stanford Sentiment Treebank text classification dataset \citep{Socher2013RecursiveDM}. For the CNN classifier, we embed the text with 50-dim GloVe vectors \cite{Pennington:2014}, feed the vectors to a ConvNet encoder, and feed the output representation into a softmax classification layer.
We use the \emph{scikit-learn} implementation of logistic regression\camready{\footnote{\url{https://scikit-learn.org}}} with bag-of-word counts and a linear classification layer.
The hyperparameter spaces $\mathcal{H}_{\text{CNN}}$ and $\mathcal{H}_{\text{LR}}$ are detailed in Appendix~\ref{app:simple_models_app}. For logistic regression we used bounds suggested by \citet{Yogatama2015BayesianOO}, which include term weighting, n-grams, stopwords, and learning rate. For the CNN we follow the hyperparameter sensitivity analysis in \citet{zhang2015}.
\resolved{; we define a reasonable neighborhoods around each hyperparameter value to search over \suchin{placeholder for better description of how we chose the search space for CNN. Honestly, it wasn't super rigorous.}}

We run 50 trials of random hyperparameter search for each classifier.
Our results (Fig.~\ref{fig:simple_models_SST (fine-grained)}) confirm previous findings \cite{zhang2015}: under a budget of fewer than 10 hyperparameter search trials, logistic regression achieves a higher expected validation accuracy than the CNN. As the budget increases, the CNN gradually improves to a higher overall expected validation accuracy. For all budgets, logistic regression has lower variance, so may be a more suitable approach for fast prototyping.
\resolved{\nascomment{talk about variance!}}

\subsection{Contextual Representations}\label{sec:contextual_embeddings}

\begin{figure}
\centering
\hspace*{-6mm}
\includegraphics[width=\linewidth]{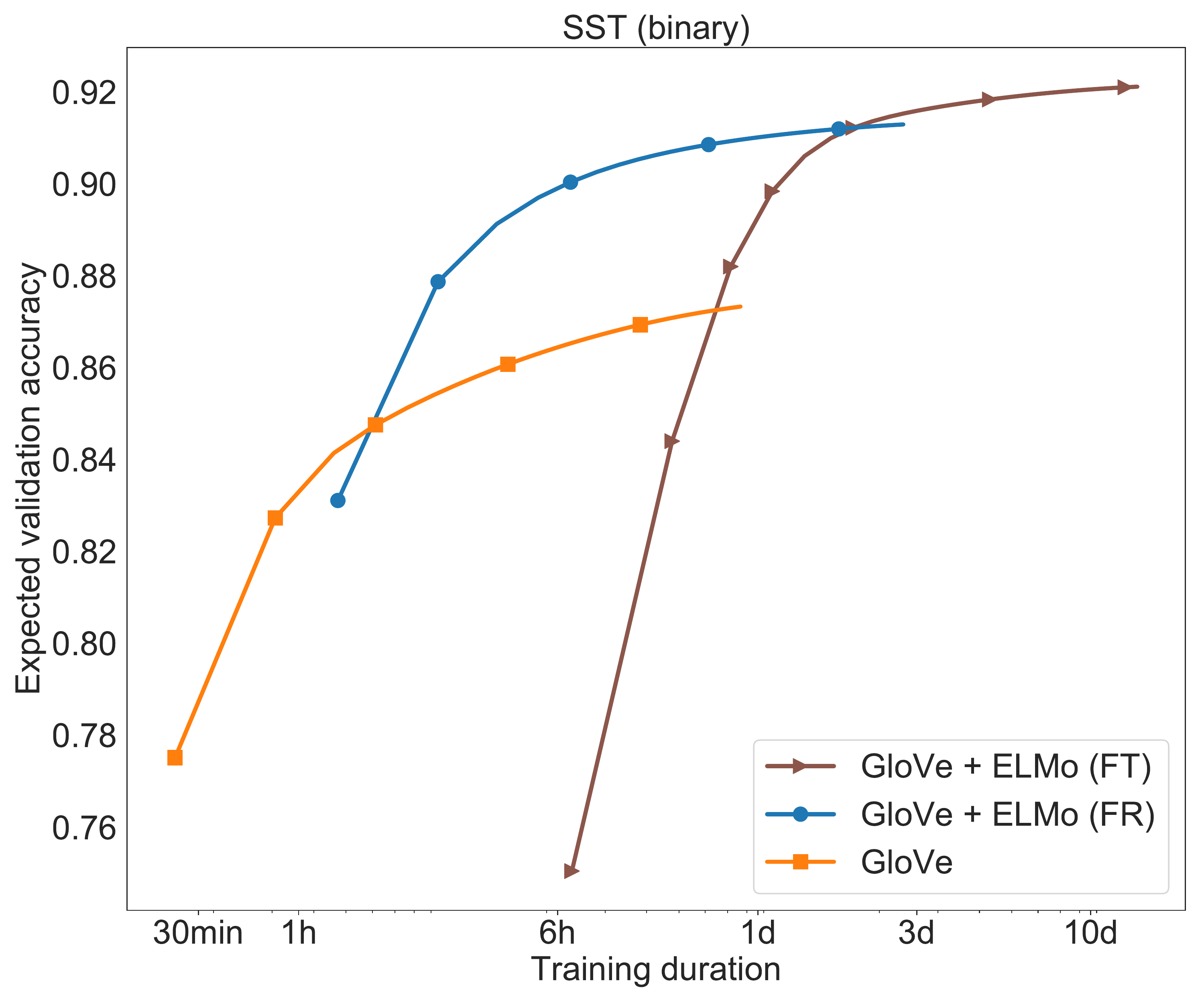}
\caption{\label{fig:bcn} Expected maximum performance of a BCN classifier on SST. We compare three embedding approaches (GloVe embeddings, GloVe + frozen ELMo, and GloVe + fine-tuned ELMo). The $x$-axis is time, on a log scale.\com{ We give the hyperparameter search space in Appendix~\ref{app:contextual_embeddings_app}.} We omit the variance for visual clarity. For each of the three model families, we sampled 50 hyperparameter values, and plot the expected maximum performance with the $x$-axis values scaled by the average training duration. 
The plot shows that for each approach (GloVe, ELMo frozen, and ELMo fine-tuned), there exists a budget for which it is preferable.
 \resolved{\dallas{This figure is a bit too wide. Can we replot with a narrower shape?}}}
\end{figure}

We next explore how computational budget affects the performance of contextual embedding models  \citep{Peters2018DeepCW}.
Recently, \citet{Peters2019ToTO} compared two methods for using contextual representations for downstream tasks: \emph{feature extraction}, where features are fixed after pretraining and passed into a task-specific model, or \emph{fine-tuning}, where they are updated during task training. 
\citet{Peters2019ToTO} found that feature extraction is preferable to fine-tuning ELMo embeddings. Here we set to explore whether this conclusion depends on the experimental budget.

Closely following their experimental setup, in Fig.~\ref{fig:bcn} we
show the expected performance of the biattentive classification network (BCN; \citealp{McCann2017LearnedIT}) with three embedding approaches (GloVe only, GloVe + ELMo frozen, and GloVe + ELMo fine-tuned), on the binary Stanford Sentiment Treebank task.\footnote{\citet{Peters2019ToTO} use a BCN with frozen embeddings and a BiLSTM BCN for fine-tuning. We conducted experiments with both a BCN and a BiLSTM with frozen and fine-tuned embeddings, and found our conclusions to be consistent. We report the full hyperparameter search space, which matched \citet{Peters2019ToTO} as closely as their reporting allowed, in Appendix \ref{app:contextual_embeddings_app}.}

We use \emph{time} for the budget by scaling the curves by the average observed training duration for each model.
We observe that as the time budget increases, the expected best-performing model changes.
\com{\citet{Peters2019ToTO} concluded that feature extraction performed the same or better than fine-tuning. 
The method used to choose hyperparameter values was not specified (likely manual search), and their search space was not fully reported, but using our setup }
In particular, we find that our experimental setup leads to the same conclusion as \citet{Peters2019ToTO} given a budget between approximately 6 hours and 1 day.
For larger budgets (e.g., 10 days) fine-tuning outperforms feature extraction.
Moreover, for smaller budgets ($< 2$ hours), using GloVe embeddings is preferable to ELMo (frozen or fine-tuned).


\subsection{Inferring Budgets in Previous Reports}\label{sec:predict_compute}
Our method provides another appealing property: estimating the budget required for the expected performance to reach a particular level, which we can compare against previously reported results.
We present two case studies, and show that the amount of computation required to match the reported results varies drastically.

We note that in the two examples that follow, the original papers only reported partial experimental information; we made sure to tune the hyperparameters they did list in addition to standard choices (such as the learning rate). 
In neither case do they report the method used to tune the hyperparameters, and we suspect they tuned them manually. 
Our experiments here are meant give an idea of the budget that would be required to reproduce their results or to apply their models to other datasets under random hyperparameter value selection. 

\begin{figure}[t!]
\includegraphics[clip,trim={0 .2cm 0 .2cm},width=\linewidth]{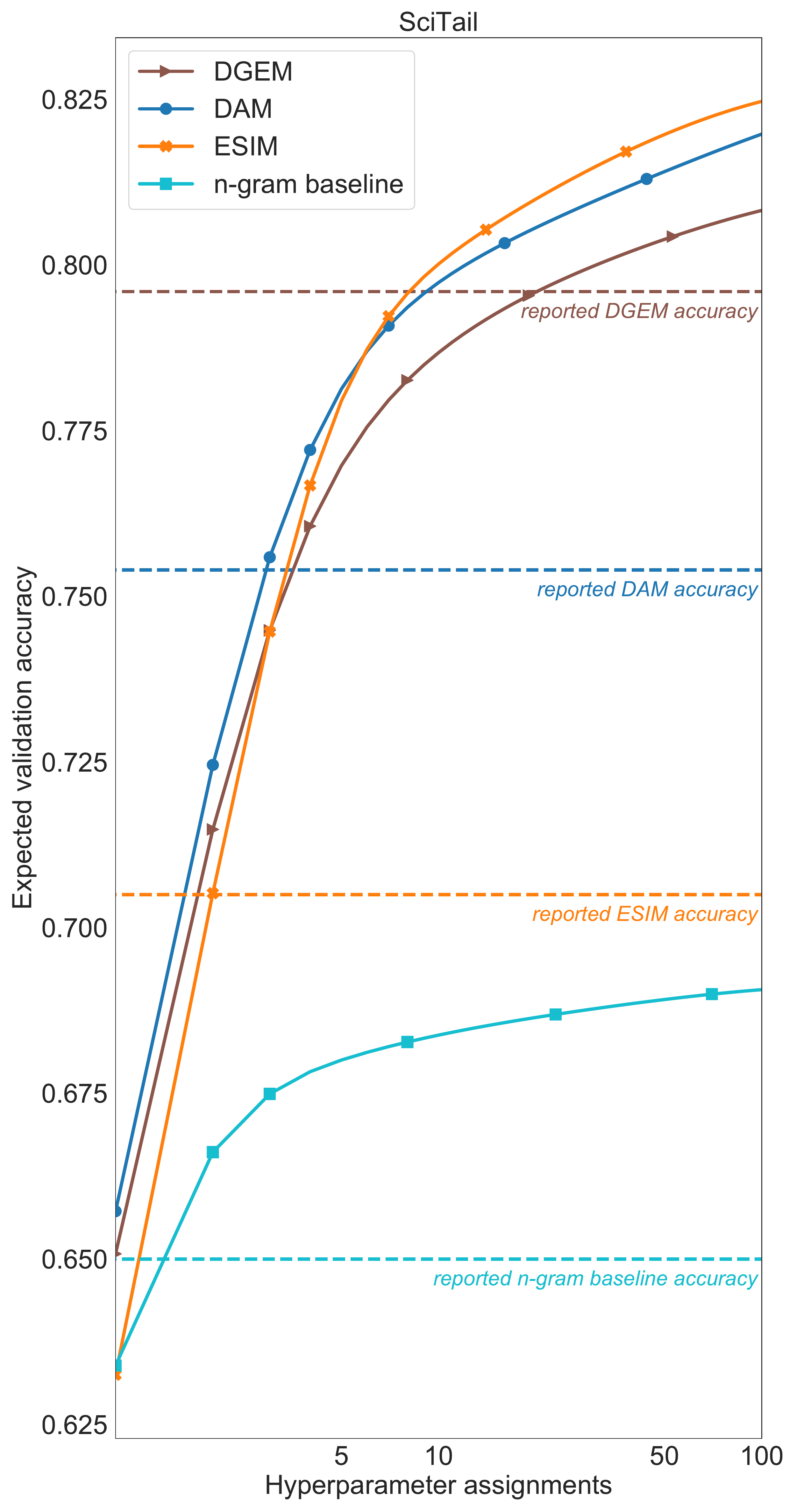}
\caption{\label{fig:scitail} Comparing reported accuracies (dashed lines) on SciTail to expected validation performance under varying levels of compute (solid lines). 
The estimated budget required for expected performance to match the reported result differs substantially across models,
and the relative ordering varies with budget.
\com{Note that while ESIM model is reported as substantially weaker model than DAM, ESIM outperforms DAM as the number of evaluated hyperparameter values grows.}
We omit variance for visual clarity.\resolved{\nascomment{explain why no s.d.?}}\resolved{ \dallas{Need to say in this figure caption that the dashed lines are what is reported in the paper!}}}
\end{figure}

\paragraph{SciTail} When introducing the SciTail textual entailment dataset, \citet{Khot2018SciTaiLAT} compared four models: an \emph{n-gram} baseline, which measures word-overlap as an indicator of entailment, \emph{ESIM} \cite{Chen2017EnhancedLF}, a sequence-based entailment model, \emph{DAM} \cite{Parikh2016ADA}, a bag-of-words entailment model, and their proposed model, \emph{DGEM} \cite{Khot2018SciTaiLAT}, a graph-based structured entailment model.
Their conclusion was that  DGEM outperforms the other models.

We use the same implementations of each of these models each with a hyperparameter search space detailed in Appendix~\ref{app:predict_compute_app}.\footnote{The search space bounds we use are large neighborhoods around the hyperparameter assignments specified in the public implementations of these models. Note that these curves depend on the specific hyperparameter search space adopted; as the original paper does not report hyperparameter search or model selection details, we have chosen what we believe to be reasonable bounds, and acknowledge that different choices could result in better or worse expected performance.}
We use a budget based on trials instead of runtime so as to emphasize how these models behave when given a comparable number of hyperparameter configurations.

Our results (Fig.~\ref{fig:scitail}) show that the different models require different budgets to reach their reported performance in expectation, ranging from 2 (n-gram) to 20 (DGEM). 
Moreover, providing a large budget for each approach improves performance substantially over reported numbers.
Finally, under different computation budgets, the top performing model changes (though the neural models are similar).

\com{
At this writing, placing in the top ten on the SciTail leaderboard\footnote{\url{leaderboard.allenai.org/scitail/}} requires performance of 80\% or better; assuming similar validation and test set performance, these results could reorder the leaderboard. We discuss leaderboards further in \S\ref{sec:leaderboards}.
\resolved{\dallas{This seems to leave something dangling.. Why haven't we tested our best model and uploaded it to the leaderboard?}}
\resolved{\dallas{Maybe we should add something like: Note that these curves depend on the specific hyperparameter search space adopted; we have chosen what we believe to be reasonable bounds, but different choices could result in slightly better or worse expected performance.}}
}

\paragraph{SQuAD} Next, we turn our attention to SQuAD \cite{Rajpurkar:2016} and report performance of the commonly-used BiDAF model \citep{Seo2017BidirectionalAF}.
The set of hyperparameters we tune covers those mentioned in addition to standard choices (details in Appendix~\ref{app:predict_compute_app}).
We see in Fig.~\ref{fig:squad} that we require a budget of 
18 GPU days
in order for the expected maximum validation performance to match the value reported in the original paper.\com{ (our full set of experiments covered 128 trials, using more than one GPU month).\roy{this comment seems irrelevant.}}
This suggests that some combination of prior intuition and extensive hyperparameter tuning were used by the original authors, though neither were reported.

\begin{figure}[t]
\hspace*{-5mm}
\includegraphics[scale=0.4]{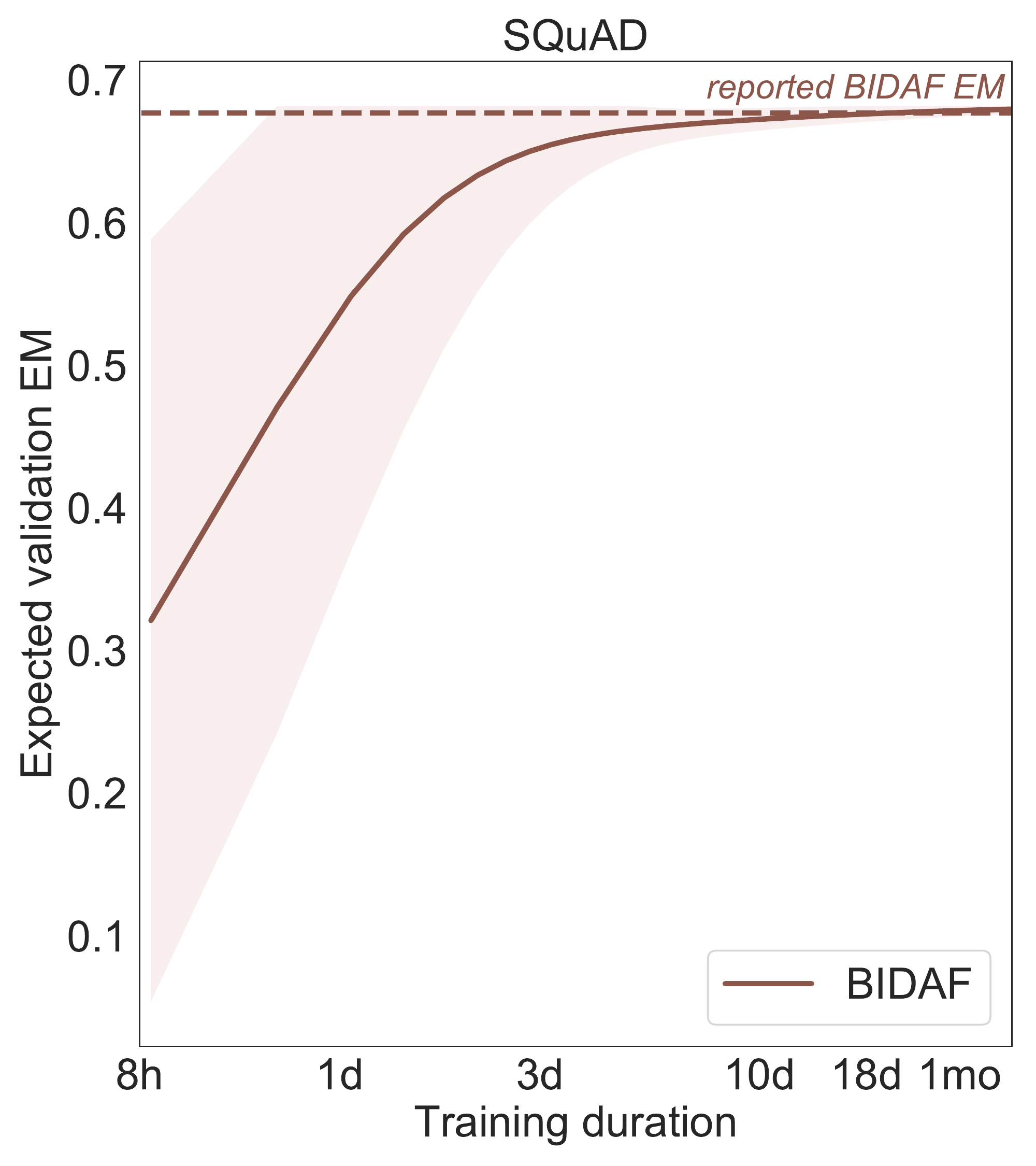}
\caption{\label{fig:squad} Comparing reported development exact-match score of BIDAF (dashed line) on SQuAD to expected performance of the best model with varying computational budgets (solid line). The shaded area represents the expected performance $\pm 1$ standard deviation, within the observed range of values.
It takes about 18 days (55 hyperparameter trials) 
for the expected performance to match the reported results.
\resolved{\nascomment{can make this figure more space-efficient by compressing y-axis or clipping at around 0.3}}}
\end{figure}


\section{Recommendations} \label{sec:recommendations}

\paragraph{Experimental results checklist}
\resolved{\nascomment{here and in caption, changed ``reporting experimental
  results checklist'' to ``experimental results checklist'' which is
  less awkward}}
The findings discussed in this paper and other similar efforts highlight methodological problems in experimental NLP. 
In this section we provide a checklist to encourage researchers to report more comprehensive experimentation results. 
Our list, shown in Text Box~\ref{tab:recommendations},\camready{mark our entries vs.~original NeurIPS list entries} builds on the reproducibility checklist that was introduced for the machine learning community during NeurIPS 2018 (which is required to be filled out for each NeurIPS 2019 submission; \citealp{reproducibility_checklist})\com{, and we recommend it for the NLP community as well}.

\newfloat{floatbox}{thp}{lob}[section]
\floatname{floatbox}{Text Box}
\counterwithout{floatbox}{section}

\begin{floatbox}[!t]
\begin{itemize}[label={\checkmark}]
\small
\item {\bf For all reported experimental results}
\begin{itemize}
    \item[$\square$] Description of computing
    infrastructure 
    \item[$\square$] Average runtime for each approach
    \item[$\square$] Details of train/validation/test splits
    \item[$\square$] Corresponding validation performance for each reported test result
    \item[$\square$] A link to implemented code
\end{itemize}
\item {\bf For experiments with hyperparameter search}
\begin{itemize}
    \item[$\square$] Bounds for each hyperparameter
    \item[$\square$] Hyperparameter configurations for best-performing models
    \item[$\square$] Number of hyperparameter search trials
    \item[$\square$] The method of choosing hyperparameter values (e.g., uniform sampling, manual tuning, etc.) and the criterion used to select among them (e.g., accuracy) \resolved{\nascomment{and the criterion used to select among them (e.g., accuracy, proxy loss, etc.) -- hopefully this won't change the next section's analysis}}
    \item[$\square$] Expected validation performance, as introduced in \S\ref{sec:expected_max}, or another measure of the mean and variance as a function of the number of hyperparameter trials.
\end{itemize}
\end{itemize}
\caption{\label{tab:recommendations} Experimental results checklist.}
\end{floatbox}

\resolved{\nascomment{do we want to add our recommendation of a plot of EVP over
  values of $n$ to this list??}}

Our focus is on improved reporting of experimental results, thus we include relevant points from their list in addition to our own.
Similar to other calls for improved reporting in machine learning \cite{model_cards,datasheets}, we recommend pairing experimental results with the information from this checklist in a structured format (see examples provided in Appendix~\ref{app:emnlp_2018_app}).

\resolved{\item \dallas{ideally, the actual dev numbers in a repo somewhere?}}

\paragraph{EMNLP 2018 checklist coverage.}\label{sec:reproducing_emnlp2018}
To estimate how commonly this information is reported in the NLP community, we sample fifty random EMNLP 2018 papers that include experimental results and evaluate how well they conform to our proposed reporting guidelines.
We find that none of the papers reported all of the items in our checklist. 
However, every paper reported at least one item in the checklist, and each item is reported by at least one paper. 
Of the papers we analyzed, 
74\% reported at least some of the best hyperparameter assignments. By contrast, 10\% or fewer papers reported hyperparameter search bounds, the number of hyperparameter evaluation trials, or measures of central tendency and variation. We include the full results of this analysis in Table~\ref{tab:emnlp} in the Appendix.

\paragraph{Comparisons with different budgets.}


\resolved{\nascomment{updated this section}}
We have argued that claims about relative model performance should be qualified by computational expense. With varying amounts of computation, not all claims about superiority are valid.
If two models have similar budgets, we can claim one outperforms the other (with that budget).
Similarly, if a model with a small budget outperforms a model with a large budget, increasing the small budget will not change this conclusion.
However, if a model with a large budget outperforms a model with a small budget, the difference might be due to the model or the budget (or both).
As a concrete example, \citet{melis2018} report the performance of an LSTM on language modeling the Penn Treebank after 1,500 rounds of Bayesian optimization; if we compare to a new $\mathcal{M}$ with a smaller budget, we can only draw a conclusion if the new model outperforms the LSTM.
\footnote{This is similar to controlling
for the amount of training data, which is an established norm in NLP research.}


In a larger sense, there may be no simple way to make a comparison ``fair.'' 
For example, the two models in Fig.~\ref{fig:simple_models_SST (fine-grained)} have hyperparameter spaces that are different, so fixing the same number of hyperparameter trials for both models does not imply a fair comparison.  
In practice, it is often not possible to measure how much past human experience has contributed to reducing the hyperparameter bounds for popular models, and there might not be a way to account for the fact that better understood (or more common) models can have better spaces to optimize over.
Further, the cost of one application of $\mathcal{A}$ might be quite different depending on the model family.  
Converting to runtime is one possible solution, but implementation effort could still affect comparisons at a fixed $x$-value.
Because of these considerations, our focus is on reporting whatever experimental results exist.
\resolved{\nascomment{please take a look at this paragraph}}

\section{Discussion: Reproducibility}\label{sec:leaderboards}

\resolved{\suchin{I think it would be helpful to include a sentence or two of how we would want leaderboards to include our approach in their setup. maybe they should include additional budget information per test/validation score? ability sort/categorize submitted models by budget? ability to compare hyperparameter search spaces? }
\jdcomment{I like these suggestions, maybe for the camera ready.}}
\resolved{\dallas{Sketching out a possible section: not sure if it belongs or not...}}

In NLP, the use of standardized test sets and public leaderboards (which limit test evaluations) has helped to mitigate the so-called ``replication crisis'' happening in fields such as psychology and medicine \citep{ioannidis.2005a,gelman.2014}. 
Unfortunately, leaderboards can create additional reproducibility issues \cite{Rogers:2019}.
First, leaderboards obscure the budget that was used to tune hyperparameters, and thus the amount of work required to apply a model to a new dataset.
Second, comparing to a model on a leaderboard is difficult if they \textit{only} report test scores.
For example, on the GLUE benchmark \cite{GLUE}, the differences in \textit{test set} performance between the top performing models can be on the order of a tenth of a percent, while the difference between test and validation performance might be one percent or larger.
\resolved{
\suchin{It would be good if we had a concrete example here, maybe the SST dataset - which we know has a large gap between test/dev? Jesse - maybe the adhoc analysis you did on SST test/dev perf would be relevant here?}
\jdcomment{this sounds like a great suggestion for the camera ready.}
}
Verifying that a new implementation matches established performance requires submitting to the leaderboard, wasting test evaluations.
Thus, we recommend leaderboards report validation performance for models evaluated on test sets.

As an example, consider \citet{BERT}, which introduced BERT and reported state-of-the-art results on the GLUE benchmark. The authors provide some details about the experimental setup, but do not report a specific budget.
Subsequent work which extended BERT \cite{stilts} included distributions of validation results, and we highlight this as a positive example of how to report experimental results.
To achieve comparable test performance to \citet{BERT}, the authors report the best of twenty or one hundred random initializations.
\resolved{\suchin{Is there a way we can quantify just how expensive this is? Maybe emphasize that training a single BERT model requires X amount of work?}
\jdcomment{That's a good idea, we should look into this for the camera ready.}}
Their validation performance reporting not only illuminates the budget required to fine-tune BERT on such tasks, but also gives other practitioners results against which they can compare without submitting to the leaderboard.
\resolved{\suchin{Should we include our plots of their results in our Appendix, and point to them here?}
\jdcomment{I think that could be nice, but i don't think we have time}}
\resolved{\suchin{We use ELMo in figure 2 - more generally, I haven't been able to find references for how long it actually takes to fine-tune BERT on one of the GLUE benchmark tasks. Only thing I've found is on SQuAD (18 hrs, according to this https://github.com/huggingface/pytorch-pretrained-BERT#examples). }
\jdcommen{hmm, we should try to find something about GLUE}
}

\section{Related Work}

\citet{lipton2018} address a number of problems with the practice of machine learning, including incorrectly attributing empirical gains to modeling choices when they came from other sources such as hyperparameter tuning.
\citet{sculley2018} list examples of similar evaluation issues, and suggest encouraging stronger standards for empirical evaluation. 
They recommend detailing experimental results found throughout the research process in a time-stamped document, as is done in other experimental science fields.
Our work formalizes these issues and provides an actionable set of recommendations to address them.

Reproducibility issues relating to standard data splits \cite{Schwartz:2011,standard_splits,  cifar_generalize, imagenet_generalize} have surfaced in a number of areas. Shuffling standard training, validation, and test set splits led to a drop in performance, and in a number of cases the inability to reproduce rankings of models.
\citet{Dror:2017} studied reproducibility in the context of consistency among multiple comparisons.

Limited community standards exist for documenting datasets and models. To address this, \citet{datasheets} recommend pairing new datasets with a ``datasheet'' which includes information such as how the data was collected, how it was cleaned, and the motivation behind building the dataset. 
Similarly, \citet{model_cards} advocate for including a ``model card'' with trained models which document training data, model assumptions, and intended use, among other things. 
Our recommendations in \S\ref{sec:recommendations} are meant to document relevant information for experimental results.






\section{Conclusion}
We have shown how current practice in experimental NLP fails to support a simple standard of reproducibility.
We introduce a new technique for estimating the expected validation performance of a method, as a function of computation budget, and present a set of recommendations for reporting experimental findings.


\section*{Acknowledgments}
This work was completed while the first author was an intern at the Allen Institute for Artificial Intelligence.
The authors thank Kevin Jamieson, Samuel Ainsworth, and the anonymous reviewers for helpful feedback.

\resolved{\roy{who gave us feedback? ARK? AI2 folks? anybody else? Noah -- funding?}}

\bibliography{emnlp-ijcnlp-2019}
\bibliographystyle{acl_natbib}

\newpage
\appendix

\onecolumn
\section{EMNLP 2018 Checklist Survey}\label{app:emnlp_2018_app}

\begin{table*}[h]
    \centering
    \begin{tabular}{cc}
       \toprule
       \textbf{Checklist item} & \textbf{Percentage of EMNLP 2018 papers}\\
       \midrule
       Reports train/validation/test splits & 92\%\\
       \midrule
       Reports best hyperparameter assignments & 74\%\\
       \midrule
       Reports code & 30\%\\
       \midrule
       Reports dev accuracy & 24\%\\
       \midrule
       Reports computing infrastructure & 18\%  \\
       \midrule
       Reports empirical runtime  & 14\% \\
       \midrule
       Reports search strategy & 14\%\\
       \midrule
       Reports score distribution & 10\%\\
       \midrule
       Reports number of hyperparameter trials & 10\%\\
       \midrule
       Reports hyperparameter search bounds & 8\%\\
       \bottomrule
    \end{tabular}
    \caption{Presence of checklist items from \S5 across 50 randomly sampled EMNLP 2018 papers that involved modeling experiments.}
    \label{tab:emnlp}
\end{table*}

    

\clearpage
\section{Hyperparameter Search Spaces for Section \ref{sec:simple_models}}\label{app:simple_models_app}

\begin{table*}[h]
    \centering
    \begin{tabular}{cc}
       \toprule
       \textbf{Computing infrastructure} & GeForce GTX 1080 GPU\\ 
       \midrule
       \textbf{Number of search trials} & 50 \\
       \midrule
       \textbf{Search strategy} & uniform sampling \\
       \midrule
       \textbf{Best validation accuracy} & 40.5\\
       \midrule
       \textbf{Training duration} & 39 sec\\
       \midrule
       \textbf{Model implementation} & \url{http://github.com/allenai/show-your-work}\\
       \bottomrule
    \end{tabular}

    \vspace{3mm}\begin{tabular}{ccc}
    \toprule
    \textbf{Hyperparameter} & \textbf{Search space} & \textbf{Best assignment} \\
    number of epochs & 50 & 50\\
    \midrule
    patience & 10 & 10\\
    \midrule
    batch size & 64 & 64\\
    \midrule
    embedding & GloVe (50 dim) & GloVe (50 dim)\\
    \midrule
    encoder & ConvNet & ConvNet\\
    \midrule
    max filter size & \emph{uniform-integer}[3, 6] & 4 \\
    \midrule
    number of filters & \emph{uniform-integer}[64, 512] & 332\\
    \midrule
    dropout & \emph{uniform-float}[0, 0.5] & 0.4\\
    \midrule
    learning rate scheduler & reduce on plateau & reduce on plateau\\
    \midrule
    learning rate scheduler patience & 2 epochs & 2 epochs\\
    \midrule
    learning rate scheduler reduction factor & 0.5 & 0.5\\
    \midrule
    learning rate optimizer & Adam & Adam\\
    \midrule
    learning rate & \emph{loguniform-float}[1e-6, 1e-1] & 0.0008 \\
    \bottomrule
    \end{tabular}
    \caption{SST (fine-grained) CNN classifier search space and best assignments.}
    \label{tab:my_label}
    
\end{table*}

\begin{table*}[h!]
\centering
\begin{tabular}{cc}
       \toprule
       \textbf{Computing Infrastructure} & 3.1 GHz Intel Core i7 CPU\\ 
       \midrule
       \textbf{Number of search trials} & 50 \\
       \midrule
       \textbf{Search strategy} & uniform sampling \\
       \midrule
       \textbf{Best validation accuracy} & 39.8\\
       \midrule
       \textbf{Training duration} & 1.56 seconds\\
       \midrule
       \textbf{Model implementation} & \url{http://github.com/allenai/show-your-work}\\
       \bottomrule
    \end{tabular}

    \vspace{3mm}\begin{tabular}{ccc}
        \toprule
        \textbf{Hyperparameter} & \textbf{Search space} & \textbf{Best assignment} \\
        \midrule
        penalty & \emph{choice}[L1, L2] & L2\\
        \midrule
        no. of iter & 100 & 100\\
        \midrule
        solver & liblinear & liblinear\\
        \midrule
        regularization & \emph{uniform-float}[0, 1] & 0.13\\
        \midrule
        n-grams & \emph{choice}[(1, 2), (1, 2, 3), (2, 3)] & [1, 2]\\
        \midrule
        stopwords & \emph{choice}[True, False] & True\\
        \midrule
        weight & \emph{choice}[tf, tf-idf, binary] & binary\\
        \midrule
        tolerance & \emph{loguniform-float}[10e-5, 10e-3] & 0.00014\\
        \bottomrule
    \end{tabular}
    \caption{SST (fine-grained) logistic regression search space and best assignments.}
    \label{tab:my_label}
\end{table*}


\clearpage

\section{Hyperparameter Search Spaces for Section \ref{sec:contextual_embeddings}}\label{app:contextual_embeddings_app}

\begin{table*}[h!]
    \centering
    \begin{tabular}{cc}
       \toprule
       \textbf{Computing Infrastructure} & GeForce GTX 1080 GPU\\ 
       \midrule
       \textbf{Number of search trials} & 50 \\
       \midrule
       \textbf{Search strategy} & uniform sampling \\
       \midrule
       \textbf{Best validation accuracy} & 87.6 \\
       \midrule
       \textbf{Training duration} & 1624 sec\\
       \midrule
       \textbf{Model implementation} & \url{http://github.com/allenai/show-your-work}\\
       \bottomrule
    \end{tabular}
    
    \vspace{3mm}\begin{tabular}{ccc}
        \toprule
        \textbf{Hyperparameter} & \textbf{Search space} & \textbf{Best assignment} \\
        \midrule
        number of epochs & 50 & 50\\
        \midrule
        patience & 10 & 10\\
        \midrule
        batch size & 64 & 64\\
        \midrule
        gradient norm & \emph{uniform-float}[5, 10] & 9.0\\
        \midrule
        embedding dropout & \emph{uniform-float}[0, 0.5] & 0.3 \\
        \midrule
        number of pre-encode feedforward layers & \emph{choice}[1, 2, 3] & 3 \\
        \midrule
        number of pre-encode feedforward hidden dims & \emph{uniform-integer}[64, 512]  & 232\\
        \midrule
        pre-encode feedforward activation & \emph{choice}[relu, tanh] & tanh\\
        \midrule
        pre-encode feedforward dropout & \emph{uniform-float}[0, 0.5]& 0.0\\
        \midrule
        encoder hidden size & \emph{uniform-integer}[64, 512] & 424 \\
        \midrule
        number of encoder layers & \emph{choice}[1, 2, 3] &  2 \\
        \midrule
        integrator hidden size &  \emph{uniform-integer}[64, 512] &  337\\
        \midrule
        number of integrator layers & \emph{choice}[1, 2, 3] & 3\\
        \midrule
        integrator dropout &  \emph{uniform-float}[0, 0.5] & 0.1\\
        \midrule
        number of output layers & \emph{choice}[1, 2, 3] & 3\\
        \midrule
        output hidden size & \emph{uniform-integer}[64, 512] & 384\\
        \midrule
        output dropout & \emph{uniform-float}[0, 0.5] & 0.2\\
        \midrule
        output pool sizes & \emph{uniform-integer}[3, 7] & 6\\
        \midrule
        learning rate optimizer & Adam & Adam\\
        \midrule
        learning rate  & \emph{loguniform-float}[1e-6, 1e-1] & 0.0001\\
        \midrule
        learning rate scheduler & reduce on plateau & reduce on plateau\\
        \midrule
        learning rate scheduler patience & 2 epochs & 2 epochs\\
        \midrule
        learning rate scheduler reduction factor & 0.5 & 0.5\\
        \bottomrule
    \end{tabular}
    \caption{SST (binary) BCN GloVe search space and best assignments.}
    \label{tab:my_label}
\end{table*}

\begin{table*}[t!]
    \centering
        \begin{tabular}{cc}
       \toprule
       \textbf{Computing Infrastructure} & GeForce GTX 1080 GPU\\ 
       \midrule
       \textbf{Number of search trials} & 50 \\
       \midrule
       \textbf{Search strategy} & uniform sampling \\
       \midrule
       \textbf{Best validation accuracy} & 91.4 \\
       \midrule
       \textbf{Training duration} & 6815 sec\\
       \midrule
       \textbf{Model implementation} & \url{http://github.com/allenai/show-your-work}\\
       \bottomrule
    \end{tabular}
    
    \vspace{3mm}\begin{tabular}{ccc}
        \toprule
        \textbf{Hyperparameter} & \textbf{Search space} & \textbf{Best assignment} \\
        \midrule
        number of epochs & 50 & 50 \\
        \midrule
        patience & 10 & 10 \\
        \midrule
        batch size & 64 & 64\\
        \midrule
        gradient norm & \emph{uniform-float}[5, 10] & 9.0\\
        \midrule
        freeze ELMo & True  & True\\ 
        \midrule
        embedding dropout & \emph{uniform-float}[0, 0.5] & 0.3\\
        \midrule
        number of pre-encode feedforward layers & \emph{choice}[1, 2, 3] & 3\\
        \midrule
        number of pre-encode feedforward hidden dims & \emph{uniform-integer}[64, 512] & 206\\
        \midrule
        pre-encode feedforward activation & \emph{choice}[relu, tanh] & relu \\
        \midrule
        pre-encode feedforward dropout & \emph{uniform-float}[0, 0.5] & 0.3\\
        \midrule
        encoder hidden size & \emph{uniform-integer}[64, 512] & 93 \\
        \midrule
        number of encoder layers & \emph{choice}[1, 2, 3] & 1 \\
        \midrule
        integrator hidden size &  \emph{uniform-integer}[64, 512] & 159\\
        \midrule
        number of integrator layers & \emph{choice}[1, 2, 3] & 3\\
        \midrule
        integrator dropout &  \emph{uniform-float}[0, 0.5] & 0.4\\
        \midrule
        number of output layers & \emph{choice}[1, 2, 3]  & 1\\
        \midrule
        output hidden size & \emph{uniform-integer}[64, 512] & 399\\
        \midrule
        output dropout & \emph{uniform-float}[0, 0.5] & 0.4\\
        \midrule
        output pool sizes & \emph{uniform-integer}[3, 7] & 6\\
        \midrule
        learning rate optimizer & Adam & Adam\\
        \midrule
        learning rate  & \emph{loguniform-float}[1e-6, 1e-1] & 0.0008\\
        \midrule
        use integrator output ELMo &   \emph{choice}[True, False] & True \\ 
        \midrule
        learning rate scheduler & reduce on plateau & reduce on plateau\\
        \midrule
        learning rate scheduler patience & 2 epochs & 2 epochs\\
        \midrule
        learning rate scheduler reduction factor & 0.5 & 0.5\\
        \bottomrule
    \end{tabular}
    \caption{SST (binary) BCN GLoVe + ELMo (frozen) search space and best assignments.}
    \label{tab:my_label}
\end{table*}

\begin{table*}[t!]
    \centering
    \begin{tabular}{cc}
       \toprule
       \textbf{Computing Infrastructure} & NVIDIA Titan Xp GPU\\ 
       \midrule
       \textbf{Number of search trials} & 50 \\
       \midrule
       \textbf{Search strategy} & uniform sampling \\
       \midrule
       \textbf{Best validation accuracy} & 92.2 \\
       \midrule
       \textbf{Training duration} & 16071 sec\\
        \midrule
       \textbf{Model implementation} & \url{http://github.com/allenai/show-your-work}\\
       \bottomrule
    \end{tabular}
    
    \vspace{3mm}\begin{tabular}{ccc}
        \toprule
        \textbf{Hyperparameter} & \textbf{Search space} & \textbf{Best assignment}\\
        \midrule
        number of epochs & 50 & 50\\
        \midrule
        patience & 10 & 10\\
        \midrule
        batch size & 64 & 64\\
        \midrule
        gradient norm & \emph{uniform-float}[5, 10] & 7.0\\
        \midrule
        freeze ELMo & False & False\\ 
        \midrule
        embedding dropout & \emph{uniform-float}[0, 0.5] & 0.1 \\
        \midrule
        number of pre-encode feedforward layers & \emph{choice}[1, 2, 3] & 3\\
        \midrule
        number of pre-encode feedforward hidden dims & \emph{uniform-integer}[64, 512] & 285 \\
        \midrule
        pre-encode feedforward activation & \emph{choice}[relu, tanh] & relu\\
        \midrule
        pre-encode feedforward dropout & \emph{uniform-float}[0, 0.5] & 0.3\\
        \midrule
        encoder hidden size & \emph{uniform-integer}[64, 512] & 368 \\
        \midrule
        number of encoder layers & \emph{choice}[1, 2, 3] & 2 \\
        \midrule
        integrator hidden size &  \emph{uniform-integer}[64, 512]  & 475\\
        \midrule
        number of integrator layers & \emph{choice}[1, 2, 3] & 3\\
        \midrule
        integrator dropout &  \emph{uniform-float}[0, 0.5] & 0.4\\
        \midrule
        number of output layers & \emph{choice}[1, 2, 3] & 3\\
        \midrule
        output hidden size & \emph{uniform-integer}[64, 512] & 362\\
        \midrule
        output dropout & \emph{uniform-float}[0, 0.5] & 0.4\\
        \midrule
        output pool sizes & \emph{uniform-integer}[3, 7] & 5\\
        \midrule
        learning rate optimizer & Adam & Adam\\
        \midrule
        learning rate  & \emph{loguniform-float}[1e-6, 1e-1] & 2.1e-5 \\
        \midrule
        use integrator output ELMo &   \emph{choice}[True, False] & True\\ 
        \midrule
        learning rate scheduler & reduce on plateau & reduce on plateau\\
        \midrule
        learning rate scheduler patience & 2 epochs & 2 epochs\\
        \midrule
        learning rate scheduler reduction factor & 0.5 & 0.5\\
        \bottomrule
    \end{tabular}
    \caption{SST (binary) BCN GloVe + ELMo (fine-tuned) search space and best assignments.}
    \label{tab:my_label}
\end{table*}

\clearpage
\section{Hyperparameter Search Spaces for Section \ref{sec:predict_compute}}\label{app:predict_compute_app}

\begin{table*}[h]
    \centering
    \begin{tabular}{cc}
       \toprule
       \textbf{Computing Infrastructure} & GeForce GTX 1080 GPU\\ 
       \midrule
       \textbf{Number of search trials} & 100 \\
       \midrule
       \textbf{Search strategy} & uniform sampling \\
       \midrule
       \textbf{Best validation accuracy} & 82.7 \\
       \midrule
       \textbf{Training duration} & 339 sec\\
       \midrule
       \textbf{Model implementation} & \url{http://github.com/allenai/show-your-work}\\
       \bottomrule
    \end{tabular}
    
    \vspace{3mm}\begin{tabular}{ccc}
        \toprule
        \textbf{Hyperparameter} & \textbf{Search space} & \textbf{Best assignment} \\
        \midrule
        number of epochs & 140 & 140 \\
        \midrule
        patience & 20 & 20 \\
        \midrule
        batch size & 64 & 64\\ 
        \midrule
        gradient clip & \emph{uniform-float}[5, 10] &  5.28\\
        \midrule
        embedding projection dim & \emph{uniform-integer}[64, 300] & 78\\
        \midrule
        number of attend feedforward layers & \emph{choice}[1, 2, 3] & 1\\
        \midrule
        attend feedforward hidden dims & \emph{uniform-integer}[64, 512] & 336 \\
        \midrule
        attend feedforward activation & \emph{choice}[relu, tanh] & tanh\\
        \midrule
        attend feedforward dropout & \emph{uniform-float}[0, 0.5] & 0.1 \\
        \midrule
        number of compare feedforward layers & \emph{choice}[1, 2, 3] & 1 \\
        \midrule
        compare feedforward hidden dims & \emph{uniform-integer}[64, 512] & 370 \\
        \midrule
        compare feedforward activation & \emph{choice}[relu, tanh]  & relu\\
        \midrule
        compare feedforward dropout & \emph{uniform-float}[0, 0.5] & 0.2\\
        \midrule
        number of aggregate feedforward layers &  \emph{choice}[1, 2, 3]  & 2\\
        \midrule
        aggregate feedforward hidden dims & \emph{uniform-integer}[64, 512] & 370\\
        \midrule
        aggregate feedforward activation & \emph{choice}[relu, tanh]& relu \\
        \midrule
        aggregate feedforward dropout & \emph{uniform-float}[0, 0.5]& 0.1\\
        \midrule
        learning rate optimizer & Adagrad & Adagrad\\
        \midrule
        learning rate & \emph{loguniform-float}[1e-6, 1e-1] & 0.009 \\
        \bottomrule
    \end{tabular}
    \caption{SciTail DAM search space and best assignments.}
    \label{tab:my_label}
\end{table*}

\begin{table*}[t!]
    \centering
    \begin{tabular}{cc}
       \toprule
       \textbf{Computing Infrastructure} & GeForce GTX 1080 GPU\\ 
       \midrule
       \textbf{Number of search trials} & 100 \\
       \midrule
       \textbf{Search strategy} & uniform sampling \\
       \midrule
       \textbf{Best validation accuracy} & 82.8 \\
       \midrule
       \textbf{Training duration} & 372 sec\\
       \midrule
       \textbf{Model implementation} & \url{http://github.com/allenai/show-your-work}\\
       \bottomrule
    \end{tabular}
    
    \vspace{3mm}\begin{tabular}{ccc}
        \toprule
        \textbf{Hyperparameter} & \textbf{Search space} & \textbf{Best assignment}  \\
        \midrule
        number of epochs & 75  & 75\\
        \midrule
        patience & 5   & 5\\
        \midrule
        batch size & 64  &  64\\
        \midrule
        encoder hidden size & \emph{uniform-integer}[64, 512]  & 253\\
        \midrule
        dropout & \emph{uniform-float}[0, 0.5]  & 0.28\\
        \midrule
        number of encoder layers & \emph{choice}[1, 2, 3]   & 1\\
        \midrule
        number of projection feedforward layers & \emph{choice}[1, 2, 3] & 2 \\ 
        \midrule
        projection feedforward hidden dims & \emph{uniform-integer}[64, 512] & 85 \\ 
        \midrule
        projection feedforward activation & \emph{choice}[relu, tanh] & relu\\
        \midrule
        number of inference encoder layers & \emph{choice}[1, 2, 3] & 1\\
        \midrule
        number of output feedforward layers & \emph{choice}[1, 2, 3] & 2 \\ 
        \midrule
        output feedforward hidden dims & \emph{uniform-integer}[64, 512] & 432 \\
        \midrule
        output feedforward activation & \emph{choice}[relu, tanh] & tanh \\
        \midrule
        output feedforward dropout & \emph{uniform-float}[0, 0.5] & 0.03 \\ 
        \midrule
        gradient norm & \emph{uniform-float}[5, 10] & 7.9 \\ 
        \midrule
        learning rate optimizer & Adam  & Adam\\
        \midrule
        learning rate & \emph{loguniform-float}[1e-6, 1e-1] & 0.0004 \\ 
        \midrule
        learning rate scheduler & reduce on plateau  & reduce on plateau \\
        \midrule
        learning rate scheduler patience & 0 epochs & 0 epochs \\
        \midrule
        learning rate scheduler reduction factor & 0.5  & 0.5 \\
        \midrule
        learning rate scheduler mode & max & max \\
        \bottomrule
    \end{tabular}
    \caption{SciTail ESIM search space and best assignments.}
    \label{tab:my_label}
\end{table*}

\begin{table*}[t!]
    \centering
    \begin{tabular}{cc}
       \toprule
       \textbf{Computing Infrastructure} & GeForce GTX 1080 GPU\\ 
       \midrule
       \textbf{Number of search trials} & 100 \\
       \midrule
       \textbf{Search strategy} & uniform sampling \\
       \midrule
       \textbf{Best validation accuracy} & 81.2 \\
       \midrule
       \textbf{Training duration} & 137 sec\\
       \midrule
       \textbf{Model implementation} & \url{http://github.com/allenai/show-your-work}\\
       \bottomrule
    \end{tabular}

    \vspace{3mm}\begin{tabular}{ccc}
        \toprule
        \textbf{Hyperparameter} & \textbf{Search space} & \textbf{Best assignment} \\
        \midrule
        number of epochs & 140 & 140 \\
        \midrule
        patience & 20 & 20 \\
        \midrule
        batch size & 64 & 64 \\
        \midrule
        dropout & \emph{uniform-float}[0, 0.5] & 0.2 \\
        \midrule
        hidden size & \emph{uniform-integer}[64, 512] & 167\\
        \midrule
        activation & \emph{choice}[relu, tanh] & tanh\\
        \midrule
        number of layers &  \emph{choice}[1, 2, 3]  & 3\\ 
        \midrule
        gradient norm & \emph{uniform-float}[5, 10] & 6.8 \\ 
        \midrule
        learning rate optimizer & Adam & Adam \\
        \midrule
        learning rate & \emph{loguniform-float}[1e-6, 1e-1] & 0.01 \\
        \midrule
        learning rate scheduler & exponential & exponential \\ 
        \midrule
        learning rate scheduler gamma & 0.5 & 0.5 \\
        \bottomrule
    \end{tabular}
    \caption{SciTail n-gram baseline search space and best assignments.}
    \label{tab:my_label}
\end{table*}

\begin{table*}[t!]
    \centering
    \begin{tabular}{cc}
       \toprule
       \textbf{Computing Infrastructure} & GeForce GTX 1080 GPU\\ 
       \midrule
       \textbf{Number of search trials} & 100 \\
       \midrule
       \textbf{Search strategy} & uniform sampling \\
       \midrule
       \textbf{Best validation accuracy} & 81.2 \\
       \midrule
       \textbf{Training duration} & 1015 sec\\
       \midrule
       \textbf{Model implementation} & \url{http://github.com/allenai/show-your-work}\\
       \bottomrule
    \end{tabular}
    
    \vspace{3mm}\begin{tabular}{ccc}
        \toprule
        \textbf{Hyperparameter} & \textbf{Search space} & \textbf{Best assignment} \\
        \midrule
        number of epochs & 140 & 140\\
        \midrule
        patience & 20 & 20\\
        \midrule
        batch size & 16 & 16\\
        \midrule
        embedding projection dim & \emph{uniform-integer}[64, 300] & 100 \\
        \midrule
        edge embedding size & \emph{uniform-integer}[64, 512] & 204  \\
        \midrule
        premise encoder hidden size & \emph{uniform-integer}[64, 512] & 234 \\
        \midrule
        number of premise encoder layers & \emph{choice}[1, 2, 3] & 2 \\
        \midrule
        premise encoder is bidirectional & \emph{choice}[True, False] & True \\
        \midrule
        number of phrase probability layers & \emph{choice}[1, 2, 3] & 2 \\
        \midrule
        phrase probability hidden dims & \emph{uniform-integer}[64, 512] & 268 \\
        \midrule
        phrase probability dropout & \emph{uniform-float}[0, 0.5] & 0.2\\
        \midrule
        phrase probability activation & \emph{choice}[tanh, relu] & tanh \\
        \midrule
        number of edge probability layers & \emph{choice}[1, 2, 3] & 1 \\
        \midrule
        edge probability dropout & \emph{uniform-float}[0, 0.5]  & 0.2\\
        \midrule
        edge probability activation & \emph{choice}[tanh, relu] & tanh \\
        \midrule
        gradient norm & \emph{uniform-float}[5, 10] & 7.0 \\
        \midrule
        learning rate optimizer & Adam & Adam\\
        \midrule
        learning rate & \emph{loguniform-float}[1e-6, 1e-1] & 0.0006 \\
        \midrule
        learning rate scheduler & exponential & exponential\\
        \midrule
        learning rate scheduler gamma & 0.5 & 0.5 \\
        \bottomrule
    \end{tabular}
    \caption{SciTail DGEM search space and best assignments.}
    \label{tab:my_label}
\end{table*}

\begin{table*}[t!]
    \centering
    \begin{tabular}{cc}
       \toprule
       \textbf{Computing Infrastructure} & GeForce GTX 1080 GPU\\ 
       \midrule
       \textbf{Number of search trials} & 128 \\
       \midrule
       \textbf{Search strategy} & uniform sampling \\
       \midrule
       \textbf{Best validation EM} & 68.2 \\
       \midrule
       \textbf{Training duration} & 31617 sec\\
       \midrule
       \textbf{Model implementation} & \url{http://github.com/allenai/show-your-work}\\
       \bottomrule
    \end{tabular}
    
    \vspace{3mm}\begin{tabular}{ccc}
        \toprule
        \textbf{Hyperparameter} & \textbf{Search space} & \textbf{Best assignment} \\
        \midrule
        number of epochs & 20 & 20 \\
        \midrule
        patience & 10 & 10\\
        \midrule
        batch size & 16 & 16 \\
        \midrule
        token embedding & GloVe (100 dim) & GloVe (100 dim)\\
        \midrule
        gradient norm & \emph{uniform-float}[5, 10] & 6.5 \\
        \midrule
        dropout & \emph{uniform-float}[0, 0.5] & 0.46 \\ 
        \midrule
        character embedding dim & \emph{uniform-integer}[16, 64] & 43  \\ 
        \midrule
        max character filter size & \emph{uniform-integer}[3, 6] & 3 \\
        \midrule 
        number of character filters & \emph{uniform-integer}[64, 512] & 33\\
        \midrule
        character embedding dropout & \emph{uniform-float}[0, 0.5] & 0.15\\
        \midrule
        number of highway layers & \emph{choice}[1, 2, 3] & 3 \\
        \midrule
        phrase layer hidden size & \emph{uniform-integer}[64, 512] & 122 \\
        \midrule
        number of phrase layers & \emph{choice}[1, 2, 3] & 1\\
        \midrule
        phrase layer dropout & \emph{uniform-float}[0, 0.5] & 0.46\\
        \midrule
        modeling layer hidden size & \emph{uniform-integer}[64, 512] & 423\\
        \midrule
        number of modeling layers & \emph{choice}[1, 2, 3] & 3\\
        \midrule
        modeling layer dropout & \emph{uniform-float}[0, 0.5] & 0.32 \\
        \midrule
        span end encoder hidden size & \emph{uniform-integer}[64, 512]  & 138\\
        \midrule
        span end encoder number of layers & \emph{choice}[1, 2, 3] & 1 \\
        \midrule
        span end encoder dropout & \emph{uniform-float}[0, 0.5] & 0.03 \\
        \midrule
        learning rate optimizer & Adam & Adam\\
        \midrule
        learning rate & \emph{loguniform-float}[1e-6, 1e-1] & 0.00056\\
        \midrule
        Adam $\beta_1$ & \emph{uniform-float}[0.9, 1.0] & 0.95\\
        \midrule
        Adam $\beta_2$ & \emph{uniform-float}[0.9, 1.0]  & 0.93\\
        \midrule
        learning rate scheduler & reduce on plateau & reduce on plateau \\
        \midrule
        learning rate scheduler patience & 2 epochs   & 2 epochs \\
        \midrule
        learning rate scheduler reduction factor  & 0.5 & 0.5 \\
        \midrule
        learning rate scheduler mode  & max & max \\
        \bottomrule
    \end{tabular}
    \caption{SQuAD BiDAF search space and best assignments.}
    \label{tab:my_label}
\end{table*}

\end{document}

\subsection{Expected max of test data}
\jdcomment{I'm not sure this fits with our framing any more, if we go with ``report dev results" as our main idea.}
While the expected max of a model on the validation set is important, we measure generalization using test data. To find the expected test performance, we again use the definition of expectations:

\begin{align} \label{eq:test_expectation}
    \E [\text{V}^t_n \mid n=\text{B}] = \sum \text{V}^t_n P(\text{V}^t_n\mid n=\text{B})
\end{align}

\roy{O didn't get the last part. why are we running things on test?}
Here we note that to get a particular test evaluation result, we evaluate $n$ hyperparameter assignments on validation data and choose the model which produced the best result $\text{V}^*_n$ to evaluate on test. Thus, the probability of getting a particular test evaluation result $\text{V}^t_n$ is equal to the probability of the best validation result $\text{V}^*_n$, and we show how to compute $P(\text{V}^*_n\mid n=B)$ in \S\ref{sec:expected_max}.

\section{comment graveyard}

\dallas{One complication is that technically I'm pretty sure we're actually dealing with a PMF here. Assuming we're working with accuracy for the case of simplicity, the number of possible values of $V_n^*$ is equal to the size of the evaluation set.}

\dallas{
\begin{itemize}
     \item It occurred to me that all of the information contained in those line plots we were thinking about (expected max performance vs computed) is actually contained in the distribution of results (without reference to time) + average runtime. In other words, if you had a numerical representation of the distribution, you can compute the line plots; However, the line plots allow for an easier comparison of at what point models differ.
     \item I think this paper will rely on really strong framing of the main idea, and on having really convincing (and surprising) results. 
     \item An important point is that (I think) most people tend to claim they have demonstrated a method is superior by getting a number that is higher than anything previously published (with the implicit assumption that the next best number tried their best to get that number, etc.). However, some of these gains have clearly come from throwing more computation at a problem. Moreover, in the same way that stopping a randomized control trial when things become significant is problematic, it is similarly problematic to keep going until you get a sufficiently good dev number and then stop searching; for proper evaluation (of a model, search space, and budget), you should then try re-running that (with comparable budgets and appropriately sized search spaces for both).
     \item I'm thinking of the above, because even some of the papers you cite at the end (like the one that proposed looking at distributions) seem to be somewhat confused about which number exactly we should be comparing.
\end{itemize}
}

\roy{Do we want to put emphasis on reporting the search space? I agree that this is important, but I am not sure I would put much focus on it because other than arguing it I don't see much we can say about it. One option is to mention it somewhere (e.g., discussion) as another key thing to report.}

\roy{Be a little more specific, the term \emph{computational budget} is not clearly defined here}.

\dallas{The convincing result would be showing some inconsistencies in previously reported results when you consider budget (similar to that other paper)}

\section{notes for jesse}

reproducing 

after meeting with kenton: reporting negative results is useful. there's a bias variance tradeoff here. where you're a practitioner, starting with the low variance things is the right first step. if there's a leaderboard, there's no way to report these lines on test.

TODO: plot biattentive and LSTM using iterations as x-axis. might be interesting, might not \jdcomment{this was not particularly interesting}. plot histograms of the validation performances.

1) he reminded me of a couple things that we should include in the paper, such as an actual histogram of performance paired with one of our expectation plots, and a comparison of a wide and narrow search space where one has higher mean and the other has higher variance.

2) we also discussed at length if we think there’s a reliable way to compare models which used different amount of compute to get results, and i feel more confident now saying it’s not, based on an argument about what the goal of reporting results is. breaking down the research process further from just “people that want the best off-the-shelf model” vs “researchers promoting a new approach”  to "people that want the best off-the-shelf model" to the three kinds of reproducibility: we want the third kind.

3) the reproducibility framing is good. 

4) when people present results, it's usually so someone else can take their conclusion about which model is best and apply it to their dataset, which is close but slightly out of domain.

Experiments Todo: 

1) fix LSTM vs biattentive, glove vs frozen vs finetuned plot so the LSTM encoder and biattentive have the same type of ELMo (LSTM ELMo). 

1.5) rerun basic model experiments on imdb.

2) NER with ELMo. not sure this is the most important. 

3) Recreating the BERT experiments in Table 7, section 5.4. This is NER, just using an LSTM (without a CRF). We likely need to find a replacement for which we can do hyperparameter search, and show in expectation how much compute was used to generate the results. This replacement should be within AllenNLP. Consider Machine Comprehension (the BiDAF implementation in AllenNLP). Another potential is DAM \url{https://arxiv.org/pdf/1606.01933.pdf}, where we would want to confirm that the AllenNLP implementation works without the EMLo embeddings, matching the results in the paper.

4) in the text classification experiments using "simple" models, we have results on SST (binary), SST (fine-grained), ag news, hatespeech, and something else (i think). we expect linear models to be the most stable, but to perform the worst eventually, while neural models are more difficult to train but eventually get better results. add more experiments until cnn overtakes boe.

a general way to inspect model performance across different tradeoff between computation and performance.

only the biggest teams can actually run these experiments. 

this approach would have been appropriate even years ago, but today 1) hyperparmeters make more of a difference, and 2) there's a larger gap between expensive models and cheap models now. 

the current work disincentivies making models faster.

Simple idea: try a narrow hyperparameter range, and then try a larger one. maybe shift it a bit (or, add dimensions which include values that weren't originally searched over, which may lead to a higher optimum). what we want is a larger variance but smaller mean for one, and a larger mean but smaller variance for the other. this would show simply how our approach can differentiate between two different approaches. 

BERT vs non-BERT: if we show this framework works for non-BERT, i think we'll be good. then, if it does or doesn't work for BERT, both are interesting results.

We want to advocate for people to report score distributions, not just point estimates. 

Hopefully we can show some things people already know: LSTMs take larger amounts of data to do well, and neural models in general require a lot of hyperparameter optimization.

Reproducibility is not just about reproducing the particular experiments we report; we can do that with Beaker. Instead, we want to present a framework that two different practitioners can use to come to the same conclusion. So, two people would both run their own experiments, and they would show that as the amount of compute increases, they see a similar max performing model.

ways this is useful:

previously, we didn't report wall time because it wasn't meaningful. however, today it does make sense for us to have some measure here of how much compute costs.

if we increase the amount of compute available, does the best model family change?

helps us understand some of the things we've seen in the research community, like melis et al., are GANs created equal?

is something people have called out: 2nd point in troubling trends in machine learning scholarship

can be useful diagnostic for someone to know if they should continue to try more hyperparameter assignments

can be useful for researchers with limited compute to understand which models might work for them, e.g. if they only have enough compute to train three models.

can be useful for researchers to compare against models trained with a ton of compute, e.g. melis et al.

cool idea: can we take a piece of existing work and use this approach to estimate how many hyperparameter assignments they tried?

make a number of vingets, where 
1) is something like showing a plot about the LSTM melis et al.
2) something about DAN, can we predict how much compute they put into the hyperparameter search? and can we show that if we had put that much compute in, we would see something that would suggest that we should try more hyperparameter assignments.
3) the lines overlap, meaning the argmax changes.

A page that estimates the max of a sample: \url{http://www.sjsu.edu/faculty/watkins/samplemax2.htm}

\section{Training algorithm}
\begin{algorithm}[h] 
 \caption{Meta-training algorithm}
 \label{alg:best_model}
 Given: Dataset: (Xtrain, Ytrain), (Xdev, Ydev), (Xtest, Ytest), budget B, hyperparameter space H, and training algorithm A, which takes in a dataset, an abstract model (M), and a hyperparameter assignment (h), and returns a trained model (C), a measure of performance on validation data (V), and time used (T): C, V, T $\leftarrow$ A(M, H, Xtrain, Ytrain, Xdev, Ydev)\;
 \KwResult{Return a single model, C*, and its performance on test data, $V^t$}
 \textbf{Initialization}:\\
 set best model C* = null\;
 set best validation performance V* = -inf\;
 \While{$B > 0$}{
  sample hyperparameters h from H: $h \sim p(h\mid H)$\;
  C, V, T $\leftarrow$ A(M, H, Xtrain, Ytrain, Xdev, Ydev)\;
  \eIf{$T \leq B$}{
   B = B - T\;
   \If{\text{V} $>$ \text{V}*} {
    V* = V\;
    C* = C\;
   }
   }{
   B = 0\;
  }
 }
 test performance $\text{V}^t$ = evaluate(C*, Xtest, Ytest)\;
 return C*, $\text{V}^t$\;
\end{algorithm}
\roy{This algorithm looks a bit trivial. do we really need it?}

\section{Alternate Derivation for Expected Max}
Let $max(\text{V}_1, \ldots, \text{V}_n) = \text{V}^*_n$. Then,

\begin{align} \label{eq:expect_as_sum_appendix}
    \E [\text{V}^*_n \mid n=\text{B}] = \sum \text{V}^*_n P(\text{V}^*_n \mid n=\text{B})
\end{align}

where, omitting the condition term for brevity, we can write out the probability of this max. For $\text{V}^*_n$ to be the max, then all $\text{V}_i$ must be less than $\text{V}_n^*$ except one, which is equal to $\text{V}_n^*$. This equality can happen in any of $n$ places, so we have

\begin{align*}
    P(\text{V}^*_n) &= P(max(\text{V}_1, \ldots, \text{V}_n) = \text{V}^*_n)\\
    &= P(\text{V}_1\leq \text{V}^*_n, \ldots, \text{V}_{n-1}\leq \text{V}^*_n, \text{V}_n= \text{V}^*_n)\\
    &+P(\text{V}_1\leq \text{V}^*_n, \ldots, \text{V}_{n-1}= \text{V}^*_n, \text{V}_n \leq \text{V}^*_n)\\
    &+\ldots\\
    &+P(\text{V}_1= \text{V}^*_n, \ldots, \text{V}_{n-1}\leq \text{V}^*_n, \text{V}_n \leq \text{V}^*_n)\\
    &=n P(\text{V}_1\leq \text{V}^*_n, \ldots, \text{V}_{n-1}\leq \text{V}^*_n, \text{V}_n= \text{V}^*_n).\\
\end{align*}

Now, using the independence of the draws, this is 

\begin{align*}
    &n P(\text{V}_1\leq \text{V}^*_n, \ldots, \text{V}_{n-1}\leq \text{V}^*_n, \text{V}_n= \text{V}^*_n)\\
    = &n P(\text{V}_1\leq \text{V}^*_n) \ldots P(\text{V}_{n-1}\leq \text{V}^*_n) P(\text{V}_n= \text{V}^*_n)\\
    = &n P(\text{V}\leq \text{V}^*_n)^{n-1} P(\text{V}= \text{V}^*_n)\\
    = &n F(\text{V})^{n-1}P(\text{V}),\\
\end{align*}

where $F(\text{V})$ is the cumulative distribution function (CDF) of V. We know the derivative of the CDF is the PDF, and we can see $n F(\text{V})^{n-1}P(\text{V}) = \frac{d}{dV} F(\text{V})^n$. Therefore, the $F(\text{V}^*_n)=F(\text{V})^n$. We can see that to compute the expectation in Eq.~\ref{eq:expect_as_sum}, we only need $F(\text{V})$. We can use the empirical CDF for this.